\documentclass[letterpaper, 10 pt, conference]{ieeeconf}  % Comment this line out if you need a4paper

\IEEEoverridecommandlockouts                              % This command is only needed if 
\usepackage{times}
\usepackage{epsfig}
\usepackage{graphicx}
\usepackage{amsmath}
\usepackage{amssymb}
\newcommand{\mypar}[1]{\vspace{0.1cm}\noindent\textbf{#1}}
\usepackage{multirow}
\usepackage{booktabs}
\usepackage[dvipsnames,table,xcdraw]{xcolor}
\usepackage{tikz}
\newcommand\copyrighttext{%
	\footnotesize Accepted at IROS 2023, \copyright IEEE. Personal use is permitted, but republication/redistribution requires IEEE permission.  Permission from IEEE must be obtained for all other uses, in any current or future media,including reprinting/republishing this material for advertising or promotional purposes, creating new collective works, for resale or redistribution to servers or lists, or reuse of any copyrighted component of this work in other works.}
\newcommand\copyrightnotice{%
	\begin{tikzpicture}[remember picture,overlay]
	\node[anchor=south,yshift=10pt] at (current page.south) {\fbox{\parbox{\dimexpr\textwidth-\fboxsep-\fboxrule\relax}{\copyrighttext}}};
	\end{tikzpicture}%
}

 \usepackage{hhline}
 \usepackage{colortbl}
 \usepackage{graphicx}

\usepackage{mathptmx} % assumes new font selection scheme installed
\usepackage{makecell}
\usepackage[T1]{fontenc}
\usepackage{bbding}
\usepackage{wasysym}

\usepackage{colortbl}
\usepackage{mathrsfs}
\usepackage{booktabs}
\usepackage{footmisc}
\usepackage{tablefootnote}
\usepackage{makecell}
\usepackage{algorithm}
\usepackage{algpseudocode}
\algnewcommand{\IIf}[1]{\State\algorithmicif\ #1\ \algorithmicthen}
\algnewcommand{\EndIIf}{\unskip\ \algorithmicend\ \algorithmicif}
\usepackage{algcompatible}

\let\labelindent\relax
\usepackage{enumitem}
\usepackage{adjustbox}
\usepackage{amssymb}% http://ctan.org/pkg/amssymb
\usepackage{pifont}% http://ctan.org/pkg/pifont
\usepackage{balance}

\usepackage{booktabs} % For professional-looking tables

\usepackage{footnote}
\makeatletter
\let\NAT@parse\undefined
\makeatother
\usepackage[colorlinks,linkcolor=red,anchorcolor=blue,urlcolor=magenta,citecolor=blue]{hyperref}

\usepackage[capitalize]{cleveref}
\crefname{section}{Sec.}{Secs.}
\Crefname{section}{Section}{Sections}
\Crefname{table}{Table}{Tables}
\crefname{table}{Tab.}{Tabs.}
\newcommand\blfootnote[1]{%
  \begingroup
  \renewcommand\thefootnote{}\footnote{#1}%
  \addtocounter{footnote}{-1}%
  \endgroup
}
\title{\LARGE \bf
Quantized Distillation: Optimizing Driver Activity Recognition Models for Resource-Constrained Environments
}
\author{Calvin Tanama$^\dag$  \quad\quad Kunyu Peng$^\dag$  \quad\quad Zdravko Marinov$^\dag$ \quad\quad  Rainer Stiefelhagen$^\dag$ \quad\quad Alina Roitberg$^{\ddag}$\vspace{0.2cm}
\\ 
\\
$^\dag$Institute for Anthropomatics and Robotics, Karlsruhe Institute of Technology \\ 
$^\ddag$Institute for Artificial Intelligence, University of Stuttgart 
\\ %\vspace{0.2cm}
}

\begin{document}

\maketitle
\copyrightnotice{}
\thispagestyle{empty}
\pagestyle{empty}

%%%%%%%%%%%%%%%%%%%%%%%%%%%%%%%%%%%%%%%%%%%%%%%%%%%%%%%%%%%%%%%%%%%%%%%%%%%%%%%%
\begin{abstract}

 Deep learning-based models are at the top of most driver observation benchmarks due to their remarkable accuracies but come with a high computational cost, while the resources are often limited in real-world driving scenarios.

This paper presents a lightweight framework for resource-efficient driver activity recognition.
We enhance 3D MobileNet, a speed-optimized neural architecture for video classification, with two paradigms for improving the trade-off between model accuracy and computational efficiency: knowledge distillation and model quantization.
Knowledge distillation prevents large drops in accuracy when reducing the model size by harvesting knowledge from a large teacher model (I3D) via soft labels instead of using the original ground truth. 
Quantization further drastically reduces the memory and computation requirements by representing the model weights and activations using lower precision integers.
Extensive experiments on a public dataset for in-vehicle monitoring during autonomous driving show that our proposed framework leads to an 3--fold reduction in model size and 1.4--fold improvement in inference time compared to an already speed-optimized architecture.
Our code is available at \url{https://github.com/calvintanama/qd-driver-activity-reco}.
\blfootnote{Emails of the authors: \tt \scriptsize \textbf{calvin.tanama}@student.kit.edu,   \{\textbf{kunyu.peng}, \textbf{zdravko.marinov}, \textbf{rainer.steifelhagen}\}@kit.edu, \textbf{alina.roitberg}@f05.uni-stuttgart.de}

\end{abstract}

%%%%%%%%%%%%%%%%%%%%%%%%%%%%%%%%%%%%%%%%%%%%%%%%%%%%%%%%%%%%%%%%%%%%%%%%%%%%%%%%
\section{Introduction}

%Why is triver activity recognition important
%There is a lot of research but still a large gap 

Efficient models for understanding the situation inside the vehicle cabin have major practical value for both manual and automated driving. 
During manual driving (SAE\footnote{Society of Automotive Engineers, Taxonomy and Definitions for Terms Related to Driving Automation Systems for On-Road Motor Vehicles:  \url{https://www.sae.org/standards/content/j3016_202104/}} levels 1-3), such models can enhance safety through identified distraction or even allow the ADAS system to foresee a dangerous maneuver by analyzing human intent, allowing the system to intervene and prevent accidents~\cite{jain2016recurrent}. 
With enhanced automation (SAE levels 4 and 5), they enable a
personalized driving experience in the case of autonomous driving, such as situation-aware adjustment of movement dynamics depending on intuitive communication via gestures.

Deep learning approaches deliver excellent results for driver activity recognition~\cite{tran2020realtime_detection_distracted, martin2019drive, tan2021bidirectional,roitberg2020cnn_spatialtemporal, roitberg2021uncertainty,zhao2021driver}, but the high computational cost of oftentimes millions of matrix multiplications required for each forward pass is their fundamental weakness and a significant bottleneck for applications within intelligent vehicles, where all sub-systems   have to share very limited computational and power resources.
One way to approach this issue is to utilize architectures based on efficient convolution variants, such as the depthwise separable convolutions within the MobileNet architecture. 
However, using such lightweight architectures comes with a considerable drop in accuracy compared to a conventional model (I3D), while the CPU inference time (2273.312 ms according to our experiments) might still not be efficient enough for very low-resource environments.  

\begin{figure}[t]
\begin{center}
\includegraphics[width = .5\textwidth]{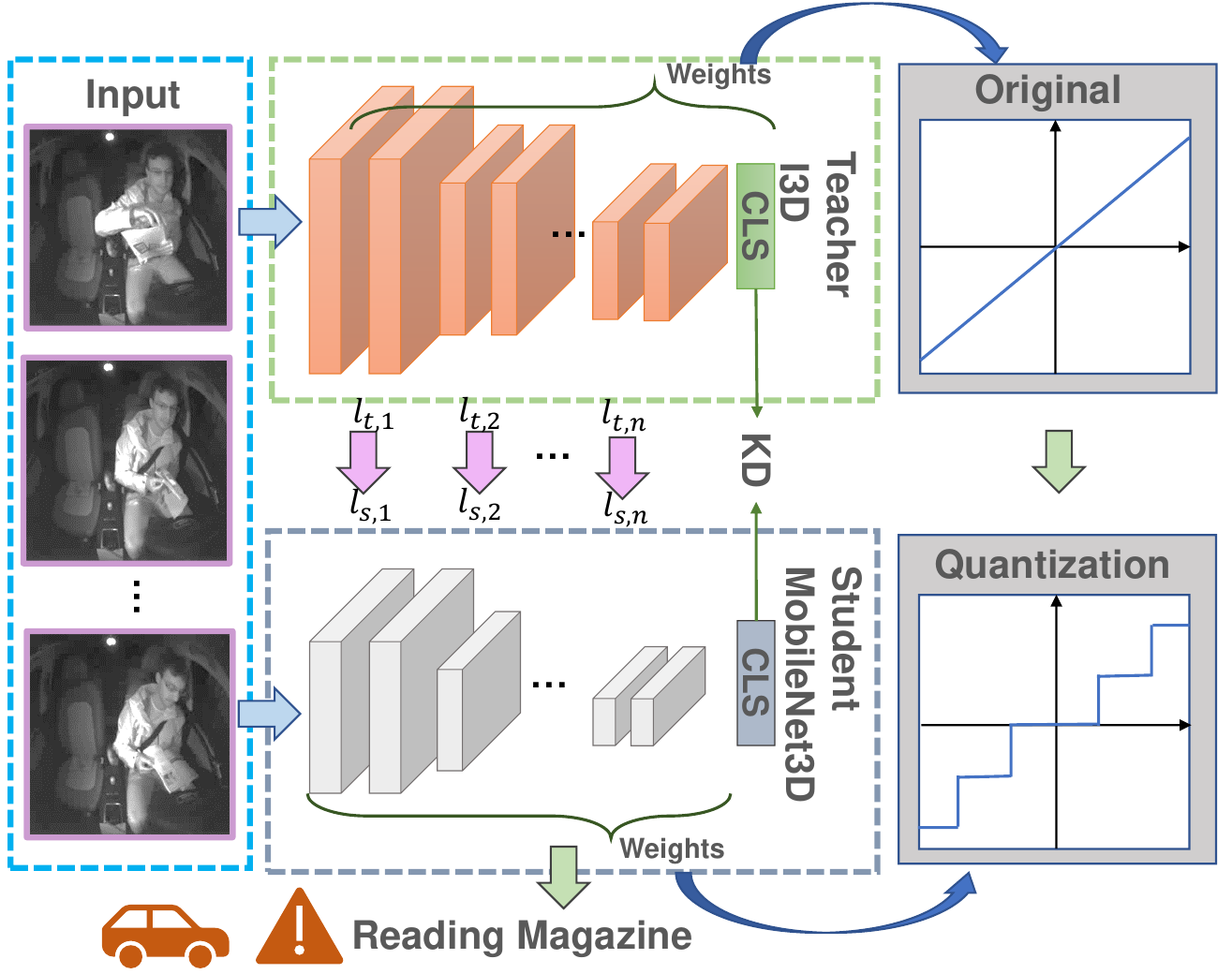}
\caption{
An overview of our proposed Quantization Distillation (QD) framework for fast driver activity recognition. QD  distills the knowledge from the logits of the teacher model (I3D) to the logits of the students model (3D MobileNet), while at the same the weights of the student network are quantized, making the architecture smaller and the inference faster. 
%We use I3D~\cite{carreira2017quo} as teacher model while using MobileNet~\cite{kopuklu2019resource} as the student model.
%Both SQD and logits-based knowledge distillation (KD) are used in our work.
}
\label{fig:first}
\end{center}
\end{figure} 

In this paper, we propose a  framework for resource-efficient driver activity recognition that addresses the trade-off between accuracy and computational efficiency. We enhance a 3D MobileNet architecture, which is a speed-optimized neural network for video classification, with two paradigms: knowledge distillation and model quantization (overview in Figure \ref{fig:first}). The former helps to prevent a large drop in accuracy when reducing the model size by transferring knowledge from a larger teacher model (I3D) via soft labels, while the latter represents the model weights and activations using lower precision integers to reduce memory and computation requirements. Our extensive experiments on a public dataset for in-vehicle monitoring during autonomous driving show that our proposed framework leads to a significant reduction in model size and inference time compared to an already speed-optimized architecture with only a slight loss in accuracy, making it suitable for deployment in resource-constrained environments. %Our work aims to guide further research in developing efficient and effective models for driver observation tasks.
%We further study the impact of different properties individual building blocks, such as the temporal footprint, width multiplier, number of bits post-quantization or knowledge distillation hyperparameters.

Overall, our work has can be summarized as the following:

\begin{itemize}
    \item We introduce a framework for resource-efficient driver activity recognition by enhancing  the speed-optimized 3D MobileNet architecture with 1) teacher-student knowledge distillation and 2) model quantization. Knowledge distillation transfers knowledge from a larger teacher model (I3D) to a smaller student model (3D MobileNet), while model quantization reduces model size and inference speed by representing model parameters with lower precision integers. 

    \item We conduct extensive experiments on the public Drive\&Act dataset for in-vehicle monitoring during autonomous driving using mean per-class accuracy, model size, and inference speed as our metrics. We also examine different properties of the individual building blocks of our models, such  as the width multiplier of MobileNet and knowledge distillation hyperparameters. 

    \item We yield a consistent reduction in model size ($3$-fold) and inference time ($1.4$-fold) compared to an already speed-optimized 3D MobileNet approach used as our backbone and comparable prediction accuracy obtained through the student-teacher training.

\end{itemize}

We hope that our work will  provide guidance for the architecture design of driver observation  models under limited resources and we will make our code publicly available.

\section{Related Work}

\noindent\textbf{Resource-Efficient Deep Learning.}
Approaches for making deep learning models more efficient can be broadly divided into three categories: arithmetic, implementation, and model-level techniques~\cite{lee2021resource}.
Arithmetic-level approaches aim to reduce the memory footprint and improve data transfer efficiency by utilizing lower precision arithmetic~\cite{xu2022mandheling, yang2022towards}. 
Implementation-level resource efficiency can be achieved through circuit optimization~\cite{7330131} and spatial architectures~\cite{7738524}.
From a model perspective, quantization-based model compression~\cite{jacob2018quantization, cheng2018model, jin2022f8net, zhang2022post, nagel2022overcoming, lin2022fq, yuan2022ptq4vit, van2022simulated, jeon2022mr, wei2022qdrop} is a well-established technique for reducing the number of bits required by the model. 
Wu \textit{et al.}\cite{wu2016quantized} and Gong \textit{et al.}\cite{gong2014compressing} leverage k-means scalar quantization to achieve resource reduction from the model parameters.
Binary model weights are used for training in BinaryNet~\cite{DBLP:journals/corr/CourbariauxB16}, BinaryConnect~\cite{courbariaux2015binaryconnect}, and XNOR networks~\cite{rastegari2016xnor}.
Wang \textit{et al.}~\cite{wang2022learnable} proposed a learnable lookup table approach for network quantization. Guo \textit{et al.}\cite{guo2022ant} exploit an adaptive numerical data type for low-bit neural network quantization. 
A different perspective has been proposed through the advent of differentiable logic gate networks~\cite{petersen2022deep}. This approach does not integrate low precision weight quantization. Instead, the structure of these architectures is exclusively built on logic operations.
Model-level resource efficiency can also be achieved through network pruning and sharing~\cite{han2015deep, hanson1988comparing, hassibi1992second, blalock2020state, rachwan2022winning, wang2022recent, li2022revisiting} and input pruning~\cite{marinov2023modselect}. Li \textit{et al.}\cite{li2022revisiting} proposed random channel pruning, while Liu \textit{et al.}\cite{liu2022soks} proposed SOKS, an automatic kernel searching-based approach for strip-wise network pruning. He \textit{et al.}~\cite{he2022sparse} introduced a sparse double dense method while considering model overfitting.
In addition to the above-mentioned techniques, knowledge distillation (KD) is a useful training strategy that can produce an efficient model by distilling knowledge from a large-scale model~\cite{hinton2015distilling, zhang2022wavelet, he2022knowledge, chen2022dearkd, liu2022transkd, zhang2022qekd}. 
Our work is inspired by the two aforementioned strategies which  we leverage to build a lightweight yet accurate driver activity recognition framework.
We enhance a 3D MobileNet architecture often used for driver activity recognition\cite{tran2020realtime_detection_distracted,li2022learning,ortega2020dmd} with  knowledge distillation and model quantization, which leads to a far better accuracy-efficiency trade-off. 
%jacob2018quantization

\noindent\textbf{Driver Activity Recognition.} Driver activity models can be divided into approaches based on manually designed feature descriptors and end-to-end deep learning approaches which operate directly on video and learn intermediate representations jointly with the classifier.
Manual feature-based methods~\cite{ohn2014head_eye_hand,xu2014realtime_random_forests,zheng2015eye_gaze,braunagel2015driver_conditionally,ohn2014head_eye_hand,braunagel2015driver_conditionally} employ classical machine learning techniques, such as Support Vector Machines and Random Forest, with features harvested from driver's hands, body- and head pose, and gaze direction.
With the advancement of deep learning techniques, end-to-end deep learning-based models have emerged as a popular approach for driver activity recognition. Convolutional neural networks (CNNs)~\cite{tran2020realtime_detection_distracted, martin2019drive, tan2021bidirectional,roitberg2020cnn_spatialtemporal, roitberg2021uncertainty,zhao2021driver, roitberg2022comparative, roitberg2022my} and transformer-based models~\cite{peng2022transdarc} are often used as backbones. 
Despite being an active research area, very few works consider inference time and model size.
The common strategy of these approaches is to leverage the speed-optimized architectures, such as MobileNet~\cite{howard2017mobilenets} as their backbones~\cite{tran2020realtime_detection_distracted,li2022learning,ortega2020dmd}.  
The work most similar to ours is presumably the recently introduced approach of Li et. al.~\cite{li2022learning}, where student-teacher distillation is leveraged to improve the lightweight MobileNet model for distracted driver posture identification.
In contrast to~\cite{li2022learning}, who propose an image-based approach, we leverage model quantization while our framework operates on  video  with a spatio-temporal CNN (3D MobileNet) as our backbone. 
A key ingredient of our approach is the quantization of model weights when training the student-teacher network to speed up the computation, which, to the best of our knowledge, has not been considered yet in the field of driver observation.

\begin{figure*}[t]
\begin{center}
\includegraphics[width = 1\textwidth]{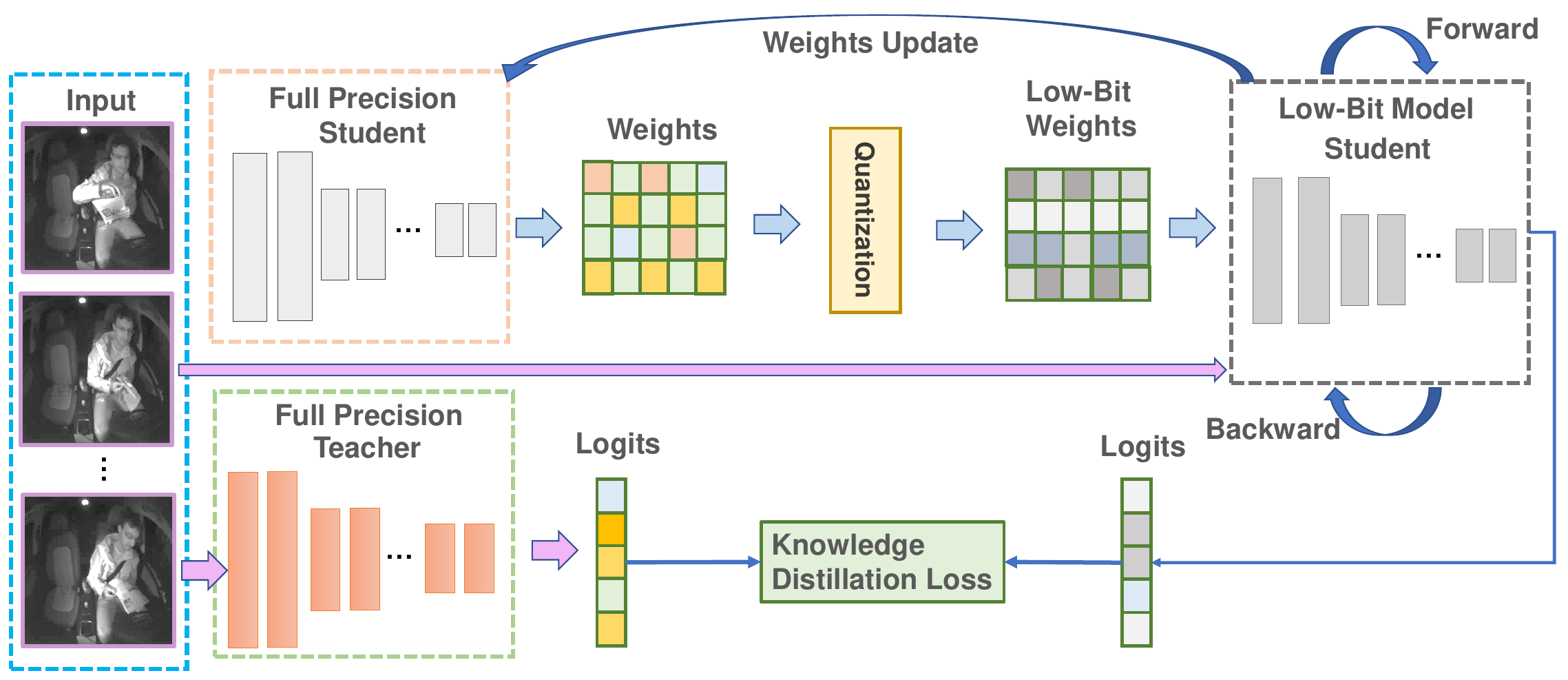}
\caption{
An overview of the Quantized Distillation training procedure with quantization-aware training. Logit vectors of the accuracy-focused teacher model (I3D) are used for loss computation and training of the lightweight student model (3D MobileNet), yielding higher accuracy than standard training with ground-truth annotations. Note, that quantization of the student weights is carried out post-training, but is also simulated during training.
}
\label{fig:main}
\end{center}
\end{figure*} 

\section{Quantized Distillation}

Our goal is to develop resource-efficient approaches for driver activity recognition. 
That is, given a video snippet $x$, our goal
is to correctly assign the potentially distractive driver behaviour, while keeping resource demands to a minimum.
We begin by employing an already speed-optimized 3D MobileNet architecture as our classification backbone.
However, our first experiments reveal two drawbacks of using this model as-is. First, the accuracy of 3D MobileNet is $\sim20\%$ below the recognition rate of an accuracy-optimized I3D architecture often used in driver observation. 
Second, despite being much more compact than I3D, even the size of $3$ MB might be too large if multiple recognition networks need to share very limited hardware resources.
To meet these challenges, we propose a framework based on two paradigms: knowledge distillation and model quantization with quantization-aware training.
Knowledge distillation mitigates a decrease in accuracy when optimizing for speed.
 This is achieved by transferring the knowledge from a larger teacher model (I3D) to a smaller student model (3D MobileNet backbone) through the use of soft labels. 
 We further map the model weights  typically represented using 32-bit floating-point numbers to lower precision integers and use quantization-aware training to reduce the amount of memory required for storage and computation.

Next, we will describe the fundamentals of the knowledge distillation mechanism (Section \ref{sec:kd}), model quantization and quantization-aware training (Section \ref{sec:quant}) used in our framework and introduce the complete Quantized Distillation approach in Section \ref{sec:quant-kd}. Finally, in Section \ref{sec:details}, we shed light on the implementation details and the used backbones.

\subsection{Knowledge Distillation}
\label{sec:kd}
Knowledge distillation (KD) is a technique for training a smaller and simpler model (the student) using a larger and more complex model (the teacher) as a guide~\cite{hinton2015distilling}. In our case, the teacher model is the I3D network \cite{carreira2017quo}, which imparts its knowledge to the smaller MobileNet3D model \cite{kopuklu2019resource}. The teacher model is trained on the labeled training data $X=\{(x_1, y_1),...(x_N, y_N)\}$ in a supervised manner, and its logits (pre-softmax outputs) for each example $x_i$ are used as "soft" labels for the student model. The student model $f_S$ is then trained to minimize a KD loss that balances a cross-entropy loss between the student's predictions and the ground truth labels with a knowledge distillation loss that encourages the student to match the teacher's "soft" labels:

\begin{equation}
\small L_{KD}(f_T, f_S, x_i) = \alpha L_{CE}(y_i, f_S(x_i)) + \beta T^2 L_{CE}\left(f_T(x_i), f_S(x_i)\right)
\label{eq:kd}
\end{equation}

Here, $\alpha$ and $\beta$ are hyperparameters that control the relative weighting of the loss terms, $T$ is a temperature scaling factor that controls the degree of smoothing applied to the predicted probabilities, and $L_{CE}$ is the cross-entropy loss. The soft labels for both teacher and student models are defined as:

\begin{equation}
f_S(x_i)_j=\frac{e^{\frac{z_s(x_i)_j}{T}}}{\sum_n e^{\frac{z_s(x_i)_n}{T}}}, \quad \text{and} \quad f_T(x_i)_j=\frac{e^{\frac{z_t(x_i)_j}{T}}}{\sum_n e^{\frac{z_t(x_i)_n}{T}}}
\end{equation}

where $z_s(x_i)_j$ and $z_t(x_i)_j$ denote the $j$-th element of the logits vector for $x_i$ computed by the student and teacher models, respectively. By minimizing the KD loss $L_{KD}$, the student model learns to produce predictions that are consistent with the teacher's predictions, while still maintaining high accuracy on the ground truth labels.

\subsection{Model Quantization}
\label{sec:quant}
Model quantization is an effective technique for reducing the memory and computational requirements of deep neural networks. It involves converting continuous values $r \in \mathbb{R}$ to a set of discrete values $q \in \mathbb{Z}$, typically by mapping floating-point numbers to integers. In practice, quantization is achieved through a \textbf{quantization scheme $(S, Z)$} that maps integers $q$ to real numbers $r$ and vice-versa:

\begin{equation}
r=S(q - Z) \text{, and } q = \text{\textbf{round}}(\frac{r}{S}-Z)
\end{equation}

where $S \in \mathbb{R}_+$ and $Z \in \mathbb{Z}$ are constant quantization parameters. The zero-point $Z$ corresponds to the quantized value of the real zero, i.e.,  $r=0$.

By translating real-number computation into quantized-value computation using the quantization scheme $(S, Z)$, inference can be performed using only integer arithmetic. We adopt the quantization scheme proposed in \cite{jacob2018quantization}, which uses the same pair of quantization parameters $(S_L, Z_L)$ for all activation and weight values within each layer $L$, but different parameters for different layers. For example, a floating-point matrix multiplication $r_3=r_2r_1$ used during a forward pass can be quantized using the corresponding parameters $S_{\alpha}, Z_{\alpha}, \alpha \in \{1,2,3\}$ as (detailed derivation in \cite{jacob2018quantization}):

\begin{equation}
q_3^{(i,k)} = Z_3 + 2^{-n} M_0 \sum_{j=1}^N(q_1^{(i,j)} - Z_1)(q_2^{(j,k)} - Z_2)
\end{equation}

Here, $q_{\alpha}^{(i,k)}$ is the quantized entry of $r_{\alpha}$ on the $i^{\text{th}}$ row and $k^{\text{th}}$ column, $M_0 \in [0.5, 1)$ is a normalized multiplier, and $n \in \mathbb{Z_+}$ is an integer. The multiplication with $M_0$ is a fixed-point multiplication, while multiplication with $2^{-n}$ is an efficient bit shift. Overall, model quantization allows us to reduce the memory footprint of deep neural networks and accelerate their inference speed, making them more suitable for deployment on resource-constrained devices. 

\noindent\textbf{Quantization Aware Training (QAT)}. During training, we leverage the QAT technique, which \textit{simulates} 8-bit quantization within the layers~\cite{jacob2018quantization}. It uses an \textit{observer} to collect statistics of input and output tensors for each layer $L$ to determine its quantization parameters $(S_L, Z_L)$. \textit{Fake quantization} layers utilize $(S_L, Z_L)$ during forward passes in training to round model parameters in each layer but keep parameters in full precision during backpropagation. QAT optimizes the model's accuracy during inference where \textbf{real} quantization is used with fewer parameter bits and integer operations. To avoid zero gradients at quantized values following a "staircase" function, a straight-through estimator (STE) \cite{esser2019learned} approximates the gradient of the \textbf{round$(\cdot)$} function with the identity function $\text{\textbf{id}}(\cdot)$, which makes it possible to train directly with quantized weights. \textit{Fused layers}  \cite{jacob2017gemmlowp} combine bias-addition, activation function evaluation, and quantized matrix multiplication into a single operation. The granularity of fused operators in inference code matches \textit{fake quantization} operators in the training.

Algorithm \ref{alg:high_level_qat_pytorch} shows the practical implementation of QAT. It involves loading a model, fusing layers to reduce rounding errors, adding observers and fake quantization, training with STE \cite{esser2019learned} for gradient computation, and 8-bit quantization of weights, batch normalization, and activations. The model can be calibrated by calculating input/output statistics in the observers. Finally, the model is converted to a quantized model for the desired back-end system.

\begin{algorithm}
\caption{Practical Implementation of QAT}
\label{alg:high_level_qat_pytorch}
\begin{algorithmic}[1]
\STATE Load pre-trained/new model
\STATE Fuse layers 
\STATE Add \textsc{Observers} and \textsc{Fake Quantization} layers 
\STATE Train with STE or calibrate model with \textsc{Observers}
\STATE Quantize the model for inference
\end{algorithmic}
\end{algorithm}

\subsection{Quantized Knowledge Distillation} 
\label{sec:quant-kd}
We propose to combine model quantization with knowledge distillation to produce an efficient 8-bit quantized model, which preserves the knowledge of a larger full-precision teacher model. To achieve this, we interleave the KD-Loss $L_{KD}$ with the QAT from Algorithm \ref{alg:high_level_qat_pytorch}. This combination can be seen in Algorithm \ref{alg:kd_with_pytorch}

\begin{algorithm}[H]
\caption{Quantized Knowledge Distillation}
\label{alg:kd_with_pytorch}
\begin{algorithmic}[1]
\STATE Let $f_S:=$  student model, $f_T:=$ teacher model, \hspace{2cm} $X=\{(x_1, y_1),...(x_N, y_N)\}:=$  training data
\STATE $f_{S} = $ {\fontfamily{qcr}\selectfont fuse\_layers($f_S$)}
\STATE $f_{S} =  $ {\fontfamily{qcr}\selectfont add\_observers\_fake\_quantization($f_S$)}
\STATE \textbf{for} $(x_i, y_i) \in X$:
    \STATE \hspace*{0.2cm} Teacher forward pass $f_T(x_i)$ 
    \STATE \hspace*{0.2cm} Student forward pass $f_S(x_i)$ \textbackslash\textbackslash \ Simulate \textbf{int8} with QAT 
    \STATE \hspace*{0.2cm} Compute distillation loss $L_{KD}(f_T, f_S, x_i)$
    \STATE \hspace*{0.21cm} Compute STE student gradient $\nabla_{f_S} = \frac{\partial L_{KD}(f_T, f_S, x_i)}{\partial f_S(x_i)}$
    \STATE \hspace*{0.2cm} Update student $f_S \leftarrow f_S - \eta \cdot \nabla_{f_S}$
\STATE $f_S \leftarrow$ {\fontfamily{qcr}\selectfont convert\_to\_int8}$(f_S)$
\STATE \textbf{return} $f_S$
\end{algorithmic}
\end{algorithm}

Algorithm \ref{alg:kd_with_pytorch} trains a student model $f_S$ using a teacher model $f_T$ to achieve similar performance to the teacher while using only 8-bit quantized weights and activations. First, bias-addition, activations, and matrix multiplications are fused into single operations in $f_S$ to simplify the architecture. Then, we add observers and fake quantization to simulate the effect of 8-bit quantization in $f_S$ during forward passes. For each training example $(x_i, y_i)$, we compute the forward pass of both the teacher and student models. We then calculate the distillation loss $L_{KD}$, which measures the discrepancy between the teacher's and student's predictions. The student gradient is calculated using the Straight-Through Estimator (STE) \cite{esser2019learned}, and the student is updated using gradient descent. After training is complete, the student model is converted to 8-bit integers and is used for efficient inference.

The proposed algorithm  produces a smaller, more efficient model that retains the knowledge of the larger, more accurate teacher model. By incorporating QAT into the knowledge distillation process, we can optimize for both model size, inference time, and accuracy simultaneously.

\subsection{Implementation Details}
\label{sec:details}

\noindent\textbf{I3D.}
Inflated 3D ConvNet (I3D)~\cite{carreira2017quo} is a widely used architecture for video analysis tasks such as action recognition. Initially, based on the Inception V2 architecture, it processes video snippets by "inflating" the original 2D architecture with an additional temporal dimension. I3D has 27 layers and approximately 25 million parameters. 

\noindent\textbf{3D MobileNet.}
The 3D MobileNet~\cite{kopuklu2019resource} architecture is a variant of the MobileNet architecture designed for 3D video analysis tasks such as action recognition. It consists of depthwise separable convolutions that greatly reduce the number of parameters and computations required by the network, making it suitable for real-time inference on mobile devices and other resource-constrained platforms. %MobileNet has 3.2 million parameters.
An important hyperparameter of MobileNet is the \textit{width multiplier} that controls the number of channels in each layer of the network (smaller width multiplier values lead to a smaller network with fewer parameters).

\noindent\textbf{Training.}  I3D is pre-trained with the Kinetics-400 dataset and trained for $200$ epochs using SGD with learning rate $0.05$ and momentum $0.9$. The learning rate is scheduled using a multi-step scheduler with a multiplier $0.2$ in epochs $\{70, 100, 150\}$. Dropout is employed with a probability of $0.5$. 3D MobileNet is pre-trained with the Kinetics-600 dataset and trained with the same SGD configuration with additional weight decay of $0.0001$ and epoch number as the baseline I3D. For Knowledge Distillation, a fully pre-trained I3D on Drive\&Act \cite{martin2019drive} is used as a teacher and MobileNet pre-trained with Kinetics-600 is used as a student. Student weight $\alpha$ is fixed to $1-\beta$ with $\beta$ being the teacher weight. Hyperparameters, \textit{e.g.}, temperature, teacher weight, and width multiplier are tuned and presented in section \ref{sec:results}. For Quantized Distillation, every \texttt{Conv-BN-ReLU} and \texttt{Conv-BN} module combination are fused. Additionally, batch normalization layers and observers are frozen on epoch $150$ and FBGEMM backend~\cite{fbgemm} is used for integer matrix multiplication. 
The experiments are implemented using PyTorch version \texttt{1.10.2}.

\noindent\textbf{Dataset and Data Processing.} Drive\&Act \cite{martin2019drive} is a public in-vehicle human activity dataset focused on distractive behavior during both, manual and autonomous driving. The data is collected from $15$ subjects  and is annotated with $34$ fine-grained activities at the main evaluation level.
We downsample the videos to a $256\times256$ resolution and use frame snippets with a size of $16$ as input to our model, as required by 3D MobileNet.
Segments of the individual activities are usually longer than that, in which case we select the frames randomly, as done in \cite{martin2019drive}. If the current segment covers fewer frames, zero-valued frames are padded in the end.
During training, we use random cropping and horizontal flipping as data augmentation and   normalize the values  to range $[0,1]$. An imbalanced data sampler is used during training to account for the uneven distribution of categories.

\begin{table}[b]
\caption{Effect of different hyperparameters on the student-teacher knowledge distillation accuracy (first validation split).}
\label{tab:hyperparameter}
\begin{center}
\begin{tabular}{  c | c | c | c  }
 \hline
 Width multiplier & $T$ & $\beta$ & Accuracy (Val 1)  \\ 
 \hline
 \textbf{0.5} & \multirow{3}{*}{5} & \multirow{3}{*}{0.7} & $\textbf{61.89\%}$ \\ 
 1.0 & & & $60.54\%$\\
 1.5 & & & $61.55\%$\\
 \hline
 \multirow{4}{*}{0.5} & 1 & \multirow{4}{*}{0.7} & $63.46\%$ \\
 & \textbf{3} & & $\textbf{63.49\%}$ \\
 & 7 & & $60.57\%$ \\
 & 9 & & $60.34\%$ \\
 \hline
 \multirow{5}{*}{0.5} & \multirow{5}{*}{3} & 0.5 & $59.05\%$\\
 & & 0.6 & $61.11\%$\\
 & & 0.8 & $63.93\%$\\
 & & \textbf{0.9} & $\textbf{64.50\%}$\\
 & & 1.0 & $62.10\%$\\
 \hline
\end{tabular}
\end{center}

\end{table}

\begin{table*}[]
\caption{Classification Accuracy and Resource-Efficiency Results}
\begin{adjustbox}{width=\linewidth,center}
\begin{tabular}{lcccccccc|ccr}
\toprule
\multicolumn{1}{c}{\multirow{2}{*}{\textbf{Method}}} & \multicolumn{8}{c}{\textbf{Accuraccy [\%]}} & \multicolumn{3}{c}{\textbf{Resource-Efficiency}} \\
\multicolumn{1}{c}{} & \multicolumn{1}{c}{\textbf{Val  1}} & \multicolumn{1}{c}{\textbf{Val  2}} & \multicolumn{1}{c}{\textbf{Val  3}} & \multicolumn{1}{c}{\textbf{Test   1}} & \multicolumn{1}{c}{\textbf{Test  2}} & \multicolumn{1}{c}{\textbf{Test  3}} & \multicolumn{1}{l}{\makecell{\textbf{Mean  (val)}}} & \multicolumn{1}{l}{\textbf{Mean (test)}} & \multicolumn{1}{c}{\textbf{Size [MB]}} & \multicolumn{1}{c}{\textbf{Time [ms]}} & \multicolumn{1}{r}{\textbf{FPS}} \\
\hline
\multicolumn{12}{l}{\cellcolor{Gray!40}Native Architectures \& Baselines} \\
\hline
Random chance & 2.94 & 2.94 & 2.94 & 2.94 & 2.94 & 2.94 & 2.94 & 2.94 & -- & -- & -- \\
I3D (accuracy-focused) & 67.04 & 63.7 & 71.31 & 60.98 & 64.93 & 53.09 & 67.35 & 59.67 & 49.41 & 703.86 & 1.42 \\
MobileNet (speed-focused) & 49.36 & 49.09 & 48.73 & 36.19 & 47.02 & 36.11 & 49.06 & 39.77 & 3.6 & 107.00 & 9.34 \\
\hline
\multicolumn{12}{l}{\cellcolor{Gray!40}Optimized Approaches (MobileNet Backbone)} \\
\hline
Knowledge Distillation & 64.5 & 58.73 & 62.01 & 49.25 & 52.11 & 44.31 & 61.75 & 48.56 & 3.6 & 107.00 & 9.34 \\
Quantization & 41.48 & 38.22 & 40.52 & 34.2 & 30.03 & 31.65 & 40.07 & 31.96 & 1.03 & 76.79 & 13.02 \\
Quantized Distillation (ours) & 59.78 & 51.48 & 61.35 & 50.11 & 50.6 & 43.79 & 57.54 & 48.17 & 1.03 & 76.79 & 13.02\\
\bottomrule
\end{tabular}
\label{tab:result}
\end{adjustbox}
\end{table*}

\begin{figure}[!t]
    \centering
    \includegraphics[width = .5\textwidth]{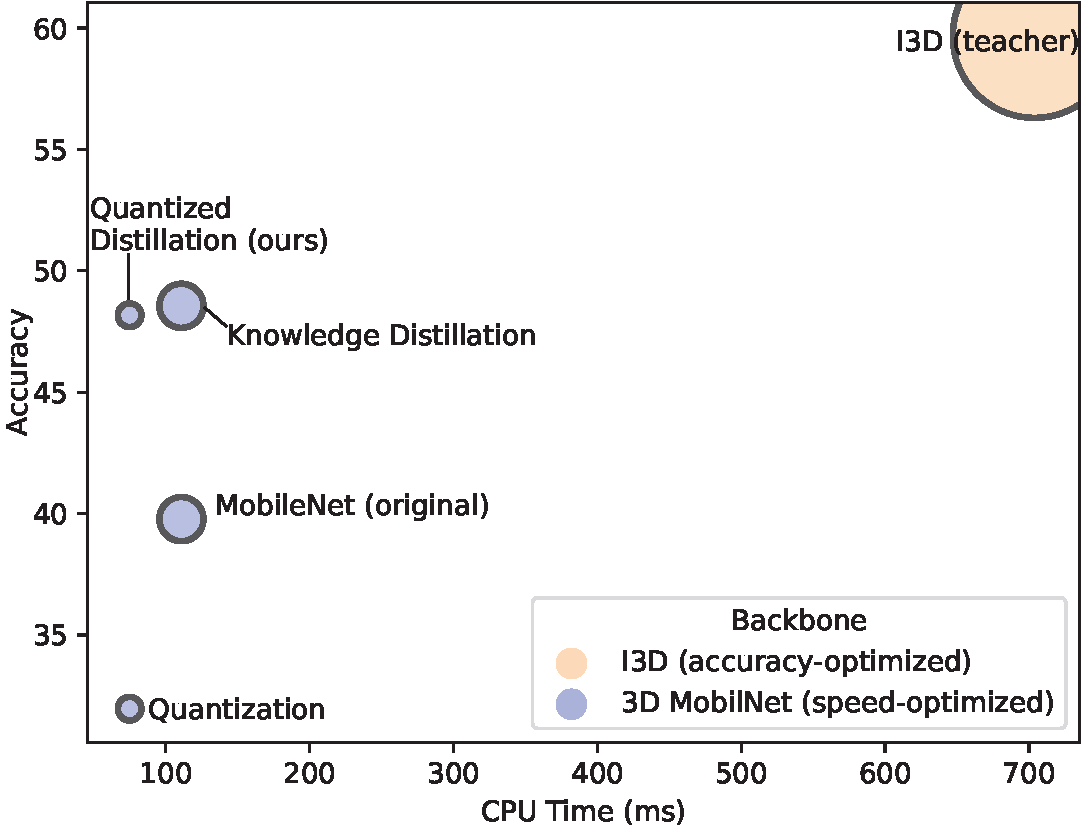}
    \caption{Comparison of model size, CPU inference time, and accuracy. Model size corresponds to the circle size. The proposed quantized distillation framework yields a good  trade-off between accuracy and computational cost.}
    \label{fig:qd_scatter}
\end{figure}

\begin{table}[]
\centering
\caption{Classification accuracies and differences for each class}
\label{tab:accuracy_class}
\begin{tabular}{l c |>{\columncolor{lightgray!80}}c| r}
\arrayrulecolor{black} % set the color of the table rules to black
\toprule
Class & MobileNet [\%] & QD  (ours) [\%] & $\Delta$ Acc \\
\midrule
closing bottle & 37.17 & 0.0 & \textcolor{Brown!70}{-37.17} \\
closing door inside & 59.89 & 63.33 & \textcolor{OliveGreen!100}{$+$3.44} \\
closing door outside & 33.33 & 88.89 & \textcolor{OliveGreen!100}{$+$55.56} \\
closing laptop & 9.52 & 26.19 & \textcolor{OliveGreen!100}{$+$16.67} \\
drinking & 22.22 & 23.87 & \textcolor{OliveGreen!100}{$+$1.65} \\
eating & 21.67 & 41.48 & \textcolor{OliveGreen!100}{$+$19.81} \\
entering car & 66.45 & 77.78 & \textcolor{OliveGreen!100}{$+$11.33} \\
exiting car & 42.33 & 71.76 & \textcolor{OliveGreen!100}{$+$29.43} \\
fastening seat belt & 28.0 & 78.79 & \textcolor{OliveGreen!100}{$+$50.79} \\
fetching an object & 21.67 & 29.72 & \textcolor{OliveGreen!100}{$+$8.06} \\
interacting with phone & 53.58 & 40.37 & \textcolor{Brown!70}{-13.22} \\
looking/moving around& 4.76 & 25.79 & \textcolor{OliveGreen!100}{$+$21.03} \\
opening backpack & 15.33 & 2.38 & \textcolor{Brown!70}{-12.95} \\
opening bottle & 12.5 & 47.44 & \textcolor{OliveGreen!100}{$+$34.94} \\
opening door inside & 68.33 & 55.0 & \textcolor{Brown!70}{-13.33} \\
opening door outside & 83.33 & 93.33 & \textcolor{OliveGreen!100}{$+$10.00} \\
opening laptop & 39.94 & 27.55 & \textcolor{Brown!70}{-12.39} \\
placing an object & 0.0 & 31.87 & \textcolor{OliveGreen!100}{$+$31.87} \\
preparing food & 0.0 & 6.67 & \textcolor{OliveGreen!100}{$+$6.67} \\
pressing autom. button & 93.94 & 94.18 & \textcolor{OliveGreen!100}{$+$0.24} \\
putting laptop in backpack & 16.67 & 0.0 & \textcolor{Brown!70}{-16.67} \\
putting on jacket & 32.29 & 0.0 & \textcolor{Brown!70}{-32.29} \\
putting on sunglasses & 23.78 & 38.89 & \textcolor{OliveGreen!100}{$+$15.11} \\
reading magazine & 43.21 & 69.82 & \textcolor{OliveGreen!100}{$+$26.61} \\
reading newspaper & 79.51 & 80.61 & \textcolor{OliveGreen!100}{$+$1.10} \\
sitting still & 58.44 & 81.41 & \textcolor{OliveGreen!100}{$+$22.97} \\
taking laptop fr. backpack & 50.00 & 25.00 & \textcolor{Brown!70}{$-$25.00} \\
taking off jacket & 46.43 & 70.05 & \textcolor{OliveGreen!100}{$+$23.62} \\
taking off sunglasses & 4.17 & 31.21 & \textcolor{OliveGreen!100}{$+$27.04} \\
talking on phone & 51.96 & 57.73 & \textcolor{OliveGreen!100}{$+$5.77} \\
unfastening seat belt & 73.07 & 64.21 & \textcolor{Brown!70}{$-$8.86} \\
using multimedia display & 94.1 & 93.2 & \textcolor{Brown!70}{$-$0.90} \\
working on laptop & 36.27 & 67.19 & \textcolor{OliveGreen!100}{$+$30.92} \\
writing & 24.91 & 32.01 & \textcolor{OliveGreen!100}{$+$7.10} \\
\bottomrule

\end{tabular}
\end{table}

\section{Experiments}
\subsection{Metrics \& Benchmark Details}

\noindent\textbf{Performance.} Following the original evaluation protocol~\cite{martin2019drive}, model recognition quality is measured by calculating the mean per class accuracy.  \\
\noindent\textbf{Model Size and Inference Time.} Model size is measured by taking the size of the saved model in the form of a \texttt{.pth} file. The inference time for one video data with $16$ frames is measured with the PyTorch \texttt{profiler} package by averaging the inference time of $1000$ forward pass. The measurement is done on an AMD EPCY 7502P 32-Core CPU.
We also mention Floating Point Operations Per Second (FLOPs) for the 3D MobileNet and I3D backbones, but do not use it in the case of the quantized model, where floating point operations are replaced with integer operations, \textit{i.e.}, the total amount of arithmetic operations stays the same, but the integer operations are executed much more efficiently. We, therefore, use the CPU inference time / Frame Per Second (FPS) and the model size as our main metrics. 

\subsection{Results}
\label{sec:results}
\noindent\textbf{Hyperparameters.} 
We first look at the effect of different hyperparameters on the student-teacher knowledge distillation outcome in Table \ref{tab:result}:  width multiplier of 3D MobileNet and  the knowledge distillation hyperparameters temperature $T$ and teacher weight $\beta$.
All hyperparameter tuning experiments are carried out on the first validation split of Drive\&Act.
To examine the width multiplier, the temperature, and teacher weight are fixed to $5$ and $0.7$ respectively. 
The width multiplier $0.5$ and $1.5$ yield comparable results. 
Since a lower width multiplier leads to a lower model size, we select $0.5$ in the next step. 
For ablations of the temperature parameter, width multiplier, and teacher weight are fixed to $0.5$ and $0.7$ respectively and models are trained with varying temperatures of $\{1,3,7,9\}$.
With the same evaluation method, temperatures $1$ and $3$ yield similar results, and the value with the best outcome, \textit{e.g.}, $3$ is taken for the next study. 
For the teacher weight, we consider values of $\{0.5, 0.6, 0.8, 0.9, 1.0\}$, where $0.9$ leads to the best outcome ($64.5\%$).  
It is worth mentioning that for all hyperparameter configurations, student-teacher knowledge distillation training consistently surpasses the original 3D MobileNet (accuracy of $49.36\%$ on the first validation split, see Table \ref{tab:result}).

\noindent\textbf{Main Results.}
Table \ref{tab:result} summarizes the accuracies and  efficiency outcomes of the developed models and baselines for the leading hyperparameter configuration (estimated on the validation set, \textit{i.e.}, width multiplier of $0.5$, $T = 3$, $\beta = 0.9$).
Looking at the original architectures, the larger I3D model outperforms 3D MobileNet by  $\sim18-20\%$ better in terms of test and validation accuracy (averaged over the three splits).
This is not surprising due to the much higher capacity of I3D ($13.75\times$ higher model size). The inference time is highly correlated to the number of FLOPs, which is $\sim373.73$ MFLOPs and $\sim27832.132$ for I3D.

Knowledge distillation with the I3D teacher raises the average accuracy of 3D MobileNet by $\sim10\%$ on both test and validation sets. As expected, there is still a gap between the average accuracy between I3D and MobileNet with knowledge distillation, as I3D has a far higher number of learnable parameters.

Quantization of 3D MobileNet weights reduces the model size  and inference time drastically: the model becomes $\sim3.5 \times$ smaller and $\sim1.4 \times$ faster. This is possible by quantizing all weights and activation values into integers and optimizing integer matrix multiplication used during inference. However, employing  quantization without any knowledge distillation reduces the accuracy by   $\sim8\%$ compared to  3D MobileNet.
Finally, our complete Quantized Distillation framework raises the recognition quality of quantized 3D MobileNet by $\sim17\%$ with $57.54\%$ and $48.17\%$ of the validation and test
examples classified correctly, surpassing the accuracy of the original MobileNet and nearly reaching the results of the non-quantized MobileNet trained with KD, yet with far better model size and inference time. 

Next, we analyze the  recognition quality for the individual driver behaviors, comparing our final Quantized Distillation (QD) framework with the original 3D MobileNet (Table \ref{tab:accuracy_class}).  
Despite being much more efficient and smaller in size, QD outperforms the original 3D MobileNet for the vast majority of behavior types.
This gain is especially large in categories \textit{closing door outside} and \textit{fastening seat belt} (around $50\%$ increase).
However, in 10 (out of 34) total categories, we observed a decline in accuracy, which was considerable for categories \textit{closing bottle} and \textit{putting on jacket} (around $30\%$ loss).
Overall, the mean per class accuracy is significantly higher for the proposed KD framework ($48.17\%$ vs.  $39.77$ on the test set, see Table \ref{tab:result}).

Figure \ref{fig:qd_scatter} summarizes the trade-off between accuracy and resource efficiency as a scatter graph, with the CPU inference time and accuracy plotted on the X and Y axes respectively, and the model size corresponding to the bubble size.
It is worth mentioning, that while the proposed QD framework meets the best accuracy-efficiency trade-off among the lightweight architectures, there is still a $10\%$ gap compared to the large I3D model. 
While this is expected, as I3D has a marge larger capacity the final model choice should be made depending on the specific requirements of the application.

\section{Conclusion}

In this work, we explored the problem of efficient driver activity recognition aimed at reducing the model size and inference time. To this end, we employed techniques such as quantization and model distillation and studied the effect of different parameters on the recognition performance. We found that these methods can significantly reduce the model size and inference time, while student-teacher knowledge distillation is useful in keeping the loss in accuracy small. In particular, we observed that a combination of quantization and distillation with carefully selected parameters can achieve the best trade-off between accuracy and efficiency. These results suggest that efficient driver activity recognition can be achieved through careful model optimization, and we hope that our study will provide guidance for the development of fast and accurate driver observation models suitable for real-life driving situations. 

\mypar{Acknowledgements.} This work was performed on the HoreKa supercomputer funded by the Ministry of Science, Research and the Arts Baden-Württemberg. Kunyu Peng was supported by the SmartAge project sponsored by the Carl Zeiss Stiftung (P2019-01-003; 2021-2026). Alina Roitberg was  supported by  the Baden-Württemberg Stiftung (Elite Postdoc Program) and Deutsche
Forschungsgemeinschaft (DFG) under Germany’s Excellence Strategy - EXC 2075.

\bibliographystyle{IEEEtran}
\balance
\bibliography{egbib}

\begin{thebibliography}{10}
\providecommand{\url}[1]{#1}
\csname url@rmstyle\endcsname
\providecommand{\newblock}{\relax}
\providecommand{\bibinfo}[2]{#2}
\providecommand\BIBentrySTDinterwordspacing{\spaceskip=0pt\relax}
\providecommand\BIBentryALTinterwordstretchfactor{4}
\providecommand\BIBentryALTinterwordspacing{\spaceskip=\fontdimen2\font plus
\BIBentryALTinterwordstretchfactor\fontdimen3\font minus
  \fontdimen4\font\relax}
\providecommand\BIBforeignlanguage[2]{{%
\expandafter\ifx\csname l@#1\endcsname\relax
\typeout{** WARNING: IEEEtran.bst: No hyphenation pattern has been}%
\typeout{** loaded for the language `#1'. Using the pattern for}%
\typeout{** the default language instead.}%
\else
\language=\csname l@#1\endcsname
\fi
#2}}

\bibitem{jain2016recurrent}
A.~Jain, A.~Singh, H.~S. Koppula, S.~Soh, and A.~Saxena, ``Recurrent neural
  networks for driver activity anticipation via sensory-fusion architecture,''
  in \emph{ICRA}.\hskip 1em plus 0.5em minus 0.4em\relax IEEE, 2016, pp.
  3118--3125.

\bibitem{tran2020realtime_detection_distracted}
D.~Tran, H.~M. Do, J.~Lu, and W.~Sheng, ``Real-time detection of distracted
  driving using dual cameras,'' in \emph{IROS}, 2020.

\bibitem{martin2019drive}
M.~Martin \emph{et~al.}, ``Drive\&act: A multi-modal dataset for fine-grained
  driver behavior recognition in autonomous vehicles,'' in \emph{ICCV}, 2019.

\bibitem{tan2021bidirectional}
M.~Tan \emph{et~al.}, ``Bidirectional posture-appearance interaction network
  for driver behavior recognition,'' \emph{T-ITS}, 2021.

\bibitem{roitberg2020cnn_spatialtemporal}
A.~Roitberg, M.~Haurilet, S.~Rei{\ss}, and R.~Stiefelhagen, ``{CNN-based}
  driver activity understanding: Shedding light on deep spatiotemporal
  representations,'' in \emph{ITSC}, 2020.

\bibitem{roitberg2021uncertainty}
A.~Roitberg, M.~Haurilet, M.~Martinez, and R.~Stiefelhagen,
  ``Uncertainty-sensitive activity recognition: A reliability benchmark and the
  {CARING} models,'' in \emph{ICPR}, 2021.

\bibitem{zhao2021driver}
L.~Zhao, F.~Yang, L.~Bu, S.~Han, G.~Zhang, and Y.~Luo, ``Driver behavior
  detection via adaptive spatial attention mechanism,'' \emph{AEI}, 2021.

\bibitem{lee2021resource}
J.~Lee, L.~Mukhanov, A.~S. Molahosseini, U.~Minhas, Y.~Hua, J.~M. del Rincon,
  K.~Dichev, C.-H. Hong, and H.~Vandierendonck, ``Resource-efficient deep
  learning: A survey on model-, arithmetic-, and implementation-level
  techniques,'' \emph{arXiv preprint arXiv:2112.15131}, 2021.

\bibitem{xu2022mandheling}
D.~Xu, M.~Xu, Q.~Wang, S.~Wang, Y.~Ma, K.~Huang, G.~Huang, X.~Jin, and X.~Liu,
  ``Mandheling: Mixed-precision on-device dnn training with dsp offloading,''
  in \emph{Annual International Conference on Mobile Computing And Networking},
  2022, pp. 214--227.

\bibitem{yang2022towards}
Y.~Yang, X.~Chi, L.~Deng, T.~Yan, F.~Gao, and G.~Li, ``Towards efficient full
  8-bit integer dnn online training on resource-limited devices without batch
  normalization,'' \emph{Neurocomputing}, 2022.

\bibitem{7330131}
C.-H. Chang, A.~S. Molahosseini, A.~A.~E. Zarandi, and T.~F. Tay, ``Residue
  number systems: A new paradigm to datapath optimization for low-power and
  high-performance digital signal processing applications,'' \emph{IEEE
  Circuits and Systems Magazine}, 2015.

\bibitem{7738524}
Y.-H. Chen, T.~Krishna, J.~S. Emer, and V.~Sze, ``Eyeriss: An energy-efficient
  reconfigurable accelerator for deep convolutional neural networks,''
  \emph{IEEE Journal of Solid-State Circuits}, vol.~52, no.~1, pp. 127--138,
  2017.

\bibitem{jacob2018quantization}
B.~Jacob, S.~Kligys, B.~Chen, M.~Zhu, M.~Tang, A.~Howard, H.~Adam, and
  D.~Kalenichenko, ``Quantization and training of neural networks for efficient
  integer-arithmetic-only inference,'' in \emph{CVPR}, 2018, pp. 2704--2713.

\bibitem{cheng2018model}
Y.~Cheng, D.~Wang, P.~Zhou, and T.~Zhang, ``Model compression and acceleration
  for deep neural networks: The principles, progress, and challenges,''
  \emph{IEEE Signal Processing Magazine}, vol.~35, no.~1, pp. 126--136, 2018.

\bibitem{jin2022f8net}
Q.~Jin, J.~Ren, R.~Zhuang, S.~Hanumante, Z.~Li, Z.~Chen, Y.~Wang, K.~Yang, and
  S.~Tulyakov, ``F8net: Fixed-point 8-bit only multiplication for network
  quantization,'' \emph{arXiv preprint arXiv:2202.05239}, 2022.

\bibitem{zhang2022post}
J.~Zhang, Y.~Zhou, and R.~Saab, ``Post-training quantization for neural
  networks with provable guarantees,'' \emph{arXiv preprint arXiv:2201.11113},
  2022.

\bibitem{nagel2022overcoming}
M.~Nagel, M.~Fournarakis, Y.~Bondarenko, and T.~Blankevoort, ``Overcoming
  oscillations in quantization-aware training,'' in \emph{ICML}.\hskip 1em plus
  0.5em minus 0.4em\relax PMLR, 2022, pp. 16\,318--16\,330.

\bibitem{lin2022fq}
Y.~Lin, T.~Zhang, P.~Sun, Z.~Li, and S.~Zhou, ``Fq-vit: Post-training
  quantization for fully quantized vision transformer,'' in \emph{IJCAI}, 2022,
  pp. 1173--1179.

\bibitem{yuan2022ptq4vit}
Z.~Yuan, C.~Xue, Y.~Chen, Q.~Wu, and G.~Sun, ``Ptq4vit: Post-training
  quantization for vision transformers with twin uniform quantization,'' in
  \emph{Computer Vision--ECCV 2022: 17th European Conference, Tel Aviv, Israel,
  October 23--27, 2022, Proceedings, Part XII}.\hskip 1em plus 0.5em minus
  0.4em\relax Springer, 2022, pp. 191--207.

\bibitem{van2022simulated}
M.~van Baalen, B.~Kahne, E.~Mahurin, A.~Kuzmin, A.~Skliar, M.~Nagel, and
  T.~Blankevoort, ``Simulated quantization, real power savings,'' in
  \emph{CVPR}, 2022, pp. 2757--2761.

\bibitem{jeon2022mr}
Y.~Jeon, C.~Lee, E.~Cho, and Y.~Ro, ``Mr. biq: Post-training non-uniform
  quantization based on minimizing the reconstruction error,'' in \emph{CVPR},
  2022, pp. 12\,329--12\,338.

\bibitem{wei2022qdrop}
X.~Wei, R.~Gong, Y.~Li, X.~Liu, and F.~Yu, ``Qdrop: randomly dropping
  quantization for extremely low-bit post-training quantization,'' \emph{arXiv
  preprint arXiv:2203.05740}, 2022.

\bibitem{wu2016quantized}
J.~Wu, C.~Leng, Y.~Wang, Q.~Hu, and J.~Cheng, ``Quantized convolutional neural
  networks for mobile devices,'' in \emph{CVPR}, 2016, pp. 4820--4828.

\bibitem{gong2014compressing}
Y.~Gong, L.~Liu, M.~Yang, and L.~Bourdev, ``Compressing deep convolutional
  networks using vector quantization,'' \emph{arXiv preprint arXiv:1412.6115},
  2014.

\bibitem{DBLP:journals/corr/CourbariauxB16}
\BIBentryALTinterwordspacing
M.~Courbariaux and Y.~Bengio, ``Binarynet: Training deep neural networks with
  weights and activations constrained to +1 or -1,'' \emph{CoRR}, vol.
  abs/1602.02830, 2016. [Online]. Available:
  \url{http://arxiv.org/abs/1602.02830}
\BIBentrySTDinterwordspacing

\bibitem{courbariaux2015binaryconnect}
M.~Courbariaux, Y.~Bengio, and J.-P. David, ``Binaryconnect: Training deep
  neural networks with binary weights during propagations,'' \emph{Advances in
  neural information processing systems}, vol.~28, 2015.

\bibitem{rastegari2016xnor}
M.~Rastegari, V.~Ordonez, J.~Redmon, and A.~Farhadi, ``Xnor-net: Imagenet
  classification using binary convolutional neural networks,'' in
  \emph{Computer Vision--ECCV 2016: 14th European Conference, Amsterdam, The
  Netherlands, October 11--14, 2016, Proceedings, Part IV}.\hskip 1em plus
  0.5em minus 0.4em\relax Springer, 2016, pp. 525--542.

\bibitem{wang2022learnable}
L.~Wang, X.~Dong, Y.~Wang, L.~Liu, W.~An, and Y.~Guo, ``Learnable lookup table
  for neural network quantization,'' in \emph{CVPR}, 2022, pp.
  12\,423--12\,433.

\bibitem{guo2022ant}
C.~Guo, C.~Zhang, J.~Leng, Z.~Liu, F.~Yang, Y.~Liu, M.~Guo, and Y.~Zhu, ``Ant:
  Exploiting adaptive numerical data type for low-bit deep neural network
  quantization,'' in \emph{2022 55th IEEE/ACM International Symposium on
  Microarchitecture (MICRO)}, 2022.

\bibitem{petersen2022deep}
F.~Petersen, C.~Borgelt, H.~Kuehne, and O.~Deussen, ``Deep differentiable logic
  gate networks,'' \emph{Advances in Neural Information Processing Systems},
  vol.~35, pp. 2006--2018, 2022.

\bibitem{han2015deep}
S.~Han, H.~Mao, and W.~J. Dally, ``Deep compression: Compressing deep neural
  networks with pruning, trained quantization and huffman coding,'' \emph{arXiv
  preprint arXiv:1510.00149}, 2015.

\bibitem{hanson1988comparing}
S.~Hanson and L.~Pratt, ``Comparing biases for minimal network construction
  with back-propagation,'' \emph{Advances in neural information processing
  systems}, vol.~1, 1988.

\bibitem{hassibi1992second}
B.~Hassibi and D.~Stork, ``Second order derivatives for network pruning:
  Optimal brain surgeon,'' \emph{Advances in neural information processing
  systems}, vol.~5, 1992.

\bibitem{blalock2020state}
D.~Blalock, J.~J. Gonzalez~Ortiz, J.~Frankle, and J.~Guttag, ``What is the
  state of neural network pruning?'' \emph{Proceedings of machine learning and
  systems}, vol.~2, pp. 129--146, 2020.

\bibitem{rachwan2022winning}
J.~Rachwan, D.~Z{\"u}gner, B.~Charpentier, S.~Geisler, M.~Ayle, and
  S.~G{\"u}nnemann, ``Winning the lottery ahead of time: Efficient early
  network pruning,'' in \emph{ICML}.\hskip 1em plus 0.5em minus 0.4em\relax
  PMLR, 2022, pp. 18\,293--18\,309.

\bibitem{wang2022recent}
H.~Wang, C.~Qin, Y.~Bai, Y.~Zhang, and Y.~Fu, ``Recent advances on neural
  network pruning at initialization,'' in \emph{IJCAI}, 2022, pp. 23--29.

\bibitem{li2022revisiting}
Y.~Li, K.~Adamczewski, W.~Li, S.~Gu, R.~Timofte, and L.~Van~Gool, ``Revisiting
  random channel pruning for neural network compression,'' in \emph{CVPR},
  2022, pp. 191--201.

\bibitem{marinov2023modselect}
Z.~Marinov, A.~Roitberg, D.~Schneider, and R.~Stiefelhagen, ``Modselect:
  Automatic modality selection for synthetic-to-real domain generalization,''
  in \emph{Computer Vision--ECCV 2022 Workshops: Tel Aviv, Israel, October
  23--27, 2022, Proceedings, Part VIII}.\hskip 1em plus 0.5em minus 0.4em\relax
  Springer, 2023, pp. 326--346.

\bibitem{liu2022soks}
G.~Liu, K.~Zhang, and M.~Lv, ``Soks: Automatic searching of the optimal kernel
  shapes for stripe-wise network pruning,'' \emph{IEEE Transactions on Neural
  Networks and Learning Systems}, 2022.

\bibitem{he2022sparse}
Z.~He, Z.~Xie, Q.~Zhu, and Z.~Qin, ``Sparse double descent: Where network
  pruning aggravates overfitting,'' in \emph{ICML}.\hskip 1em plus 0.5em minus
  0.4em\relax PMLR, 2022, pp. 8635--8659.

\bibitem{hinton2015distilling}
G.~Hinton, O.~Vinyals, and J.~Dean, ``Distilling the knowledge in a neural
  network,'' \emph{arXiv preprint arXiv:1503.02531}, 2015.

\bibitem{zhang2022wavelet}
L.~Zhang, X.~Chen, X.~Tu, P.~Wan, N.~Xu, and K.~Ma, ``Wavelet knowledge
  distillation: Towards efficient image-to-image translation,'' in \emph{CVPR},
  2022, pp. 12\,464--12\,474.

\bibitem{he2022knowledge}
R.~He, S.~Sun, J.~Yang, S.~Bai, and X.~Qi, ``Knowledge distillation as
  efficient pre-training: Faster convergence, higher data-efficiency, and
  better transferability,'' in \emph{Proceedings of the IEEE/CVF Conference on
  Computer Vision and Pattern Recognition}, 2022, pp. 9161--9171.

\bibitem{chen2022dearkd}
X.~Chen, Q.~Cao, Y.~Zhong, J.~Zhang, S.~Gao, and D.~Tao, ``Dearkd:
  data-efficient early knowledge distillation for vision transformers,'' in
  \emph{CVPR}, 2022, pp. 12\,052--12\,062.

\bibitem{liu2022transkd}
R.~Liu, K.~Yang, A.~Roitberg, J.~Zhang, K.~Peng, H.~Liu, and R.~Stiefelhagen,
  ``Transkd: Transformer knowledge distillation for efficient semantic
  segmentation,'' \emph{arXiv preprint arXiv:2202.13393}, 2022.

\bibitem{zhang2022qekd}
J.~Zhang, C.~Chen, J.~Dong, R.~Jia, and L.~Lyu, ``Qekd: query-efficient and
  data-free knowledge distillation from black-box models,'' \emph{arXiv
  preprint arXiv:2205.11158}, 2022.

\bibitem{li2022learning}
W.~Li, J.~Wang, T.~Ren, F.~Li, J.~Zhang, and Z.~Wu, ``Learning accurate,
  speedy, lightweight cnns via instance-specific multi-teacher knowledge
  distillation for distracted driver posture identification,'' \emph{IEEE
  Transactions on Intelligent Transportation Systems}, vol.~23, no.~10, pp.
  17\,922--17\,935, 2022.

\bibitem{ortega2020dmd}
J.~D. Ortega, N.~Kose, P.~Ca{\~n}as, M.-A. Chao, A.~Unnervik, M.~Nieto,
  O.~Otaegui, and L.~Salgado, ``Dmd: A large-scale multi-modal driver
  monitoring dataset for attention and alertness analysis,'' in \emph{Computer
  Vision--ECCV 2020 Workshops: Glasgow, UK, August 23--28, 2020, Proceedings,
  Part IV 16}.\hskip 1em plus 0.5em minus 0.4em\relax Springer, 2020, pp.
  387--405.

\bibitem{ohn2014head_eye_hand}
E.~Ohn-Bar, S.~Martin, A.~Tawari, and M.~M. Trivedi, ``Head, eye, and hand
  patterns for driver activity recognition,'' in \emph{ICPR}, 2014.

\bibitem{xu2014realtime_random_forests}
L.~Xu and K.~Fujimura, ``Real-time driver activity recognition with random
  forests,'' in \emph{AutomotiveUI}, 2014.

\bibitem{zheng2015eye_gaze}
R.~Zheng, K.~Nakano, H.~Ishiko, K.~Hagita, M.~Kihira, and T.~Yokozeki,
  ``Eye-gaze tracking analysis of driver behavior while interacting with
  navigation systems in an urban area,'' \emph{THMS}, 2016.

\bibitem{braunagel2015driver_conditionally}
C.~Braunagel, E.~Kasneci, W.~Stolzmann, and W.~Rosenstiel, ``Driver-activity
  recognition in the context of conditionally autonomous driving,'' in
  \emph{ITSC}, 2015.

\bibitem{roitberg2022comparative}
A.~Roitberg, K.~Peng, Z.~Marinov, C.~Seibold, D.~Schneider, and
  R.~Stiefelhagen, ``A comparative analysis of decision-level fusion for
  multimodal driver behaviour understanding,'' in \emph{2022 IEEE Intelligent
  Vehicles Symposium (IV)}.\hskip 1em plus 0.5em minus 0.4em\relax IEEE, 2022,
  pp. 1438--1444.

\bibitem{roitberg2022my}
A.~Roitberg, K.~Peng, D.~Schneider, K.~Yang, M.~Koulakis, M.~Martinez, and
  R.~Stiefelhagen, ``Is my driver observation model overconfident? input-guided
  calibration networks for reliable and interpretable confidence estimates,''
  \emph{IEEE Transactions on Intelligent Transportation Systems}, vol.~23,
  no.~12, pp. 25\,271--25\,286, 2022.

\bibitem{peng2022transdarc}
K.~Peng, A.~Roitberg, K.~Yang, J.~Zhang, and R.~Stiefelhagen, ``Transdarc:
  Transformer-based driver activity recognition with latent space feature
  calibration,'' in \emph{IROS}.\hskip 1em plus 0.5em minus 0.4em\relax IEEE,
  2022, pp. 278--285.

\bibitem{howard2017mobilenets}
A.~G. Howard, M.~Zhu, B.~Chen, D.~Kalenichenko, W.~Wang, T.~Weyand,
  M.~Andreetto, and H.~Adam, ``Mobilenets: Efficient convolutional neural
  networks for mobile vision applications,'' \emph{arXiv preprint
  arXiv:1704.04861}, 2017.

\bibitem{carreira2017quo}
J.~Carreira and A.~Zisserman, ``Quo vadis, action recognition? {A} new model
  and the kinetics dataset,'' in \emph{CVPR}, 2017.

\bibitem{kopuklu2019resource}
O.~Kopuklu, N.~Kose, A.~Gunduz, and G.~Rigoll, ``Resource efficient 3d
  convolutional neural networks,'' in \emph{ICCV Workshops}, 2019.

\bibitem{esser2019learned}
S.~K. Esser, J.~L. McKinstry, D.~Bablani, R.~Appuswamy, and D.~S. Modha,
  ``Learned step size quantization,'' \emph{arXiv preprint arXiv:1902.08153},
  2019.

\bibitem{jacob2017gemmlowp}
B.~Jacob and P.~Warden, ``gemmlowp: A small self-contained low-precision gemm
  library,'' \emph{Retrieved June}, vol.~14, p. 2018, 2017.

\bibitem{fbgemm}
D.~Khudia, J.~Huang, P.~Basu, S.~Deng, H.~Liu, J.~Park, and M.~Smelyanskiy,
  ``Fbgemm: Enabling high-performance low-precision deep learning inference,''
  \emph{arXiv preprint arXiv:2101.05615}, 2021.

\end{thebibliography}
%\balance

\end{document}